\documentclass{article}

\usepackage[preprint]{neurips_2026}

\usepackage[utf8]{inputenc}
\usepackage[T1]{fontenc}
\usepackage{hyperref}
\hypersetup{
    colorlinks=true,
    citecolor=teal,      
    linkcolor=black,     
    urlcolor=teal,
    filecolor=magenta
}
\usepackage{url}
\usepackage{booktabs}
\usepackage{amsfonts}
\usepackage{amssymb}
\usepackage{pifont}
\usepackage{graphicx}
\usepackage{algorithm}
\usepackage{algpseudocode}
\usepackage{float}
\usepackage{wrapfig}
\usepackage{makecell}
\usepackage{listings}      
\usepackage{nicefrac}
\usepackage{microtype}
\usepackage[table]{xcolor}
\usepackage{amsmath}
\usepackage[nameinlink,capitalise,noabbrev]{cleveref}
\usepackage{multirow}
\usepackage{subcaption}
\usepackage{tcolorbox}
\tcbuselibrary{skins}

\usepackage{array}
\usepackage{ragged2e}      
\usepackage{fontawesome5}  
\usepackage{siunitx}       
\definecolor{skyblue}{HTML}{92C5DE}
\definecolor{myblue}{HTML}{6691CD}
\definecolor{hl}{RGB}{220,235,255}   

\definecolor{cvpdrow}{RGB}{238,246,255}

\newtcolorbox{HighlighterBox}[2][]{
	arc=3.8pt,
	left=5.0pt,
	right=5.0pt,
	bottom=2pt,
	top=2pt,
	colback=skyblue!7.5,
	colframe=skyblue!35,
	boxrule=0.8pt,
	colbacktitle=skyblue!35,
	coltitle=myblue!20!black,
	title=\textbf{#2},
	fonttitle=\bfseries,
	before upper=\justifying,
	#1,
}

\definecolor{ncmetabg}{HTML}{F1F4F7}
\definecolor{ncmetaedge}{HTML}{DCE6F5}
\definecolor{ncmetablue}{HTML}{1877F2}
\definecolor{ncapricot}{HTML}{F7EBDD}

\usepackage{xcolor}
\newif\ifshowcolors
\showcolorsfalse

\ifshowcolors

\else

\fi

\title{%
    \begin{minipage}[c]{0.12\textwidth}
        \centering
        \includegraphics[height=1.3cm,keepaspectratio]{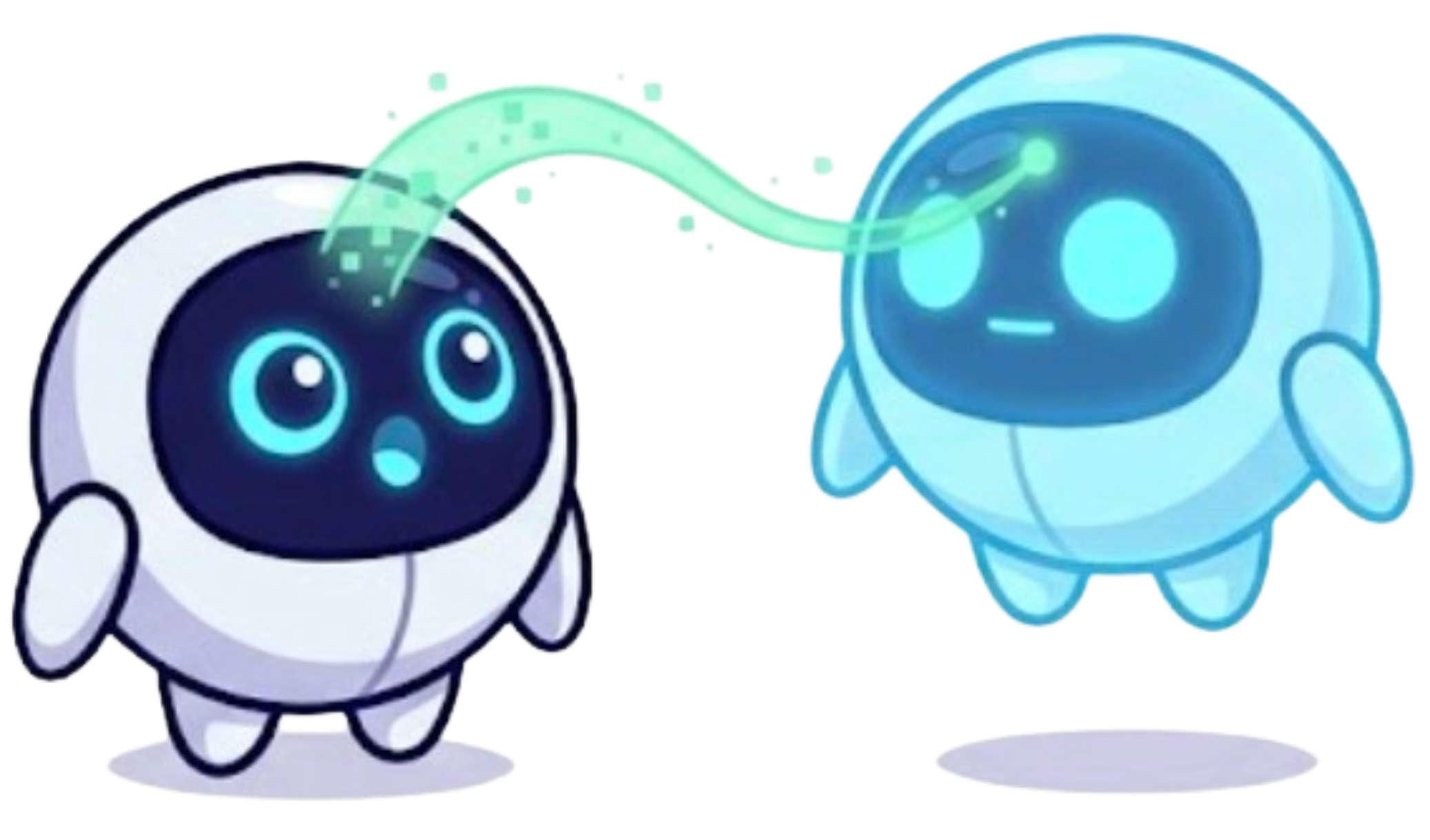}
    \end{minipage}%
    \hspace{0.02\textwidth}%
    \begin{minipage}[c]{0.9\textwidth}
        \centering
        \textbf{Perception Before Supervision: Self-Contained Visual Distillation from Counterfactual Blind Spots}
    \end{minipage}%
}

\author{%
  \normalfont
  \textbf{Shravan Venkatraman}$^{1}$\thanks{Correspondence: \texttt{shravan.venkatraman@mbzuai.ac.ae}}, \textbf{Omkar Thawakar}$^{1}$, \textbf{Ritesh Thawkar}$^{1}$, \\ \textbf{Abdelrahman Shaker}$^{1}$, \textbf{Rao Muhammad Anwer}$^{1,2}$ \\[2pt]
  $^{1}$Mohamed bin Zayed University of Artificial Intelligence \quad
  $^{2}$Aalto University \\
}
\makeatletter
\newcommand{\maketitleboxed}[1]{%
  \par
  \begingroup
    \renewcommand{\thefootnote}{\fnsymbol{footnote}}
    \renewcommand{\@makefnmark}{\textsuperscript{\@thefnmark}}
    \long\def\@makefntext##1{%
      \parindent 1em\noindent
      \hbox to 1.8em{\hss $\m@th ^{\@thefnmark}$}##1%
    }
    \thispagestyle{empty}%
    \begin{tcolorbox}[
      enhanced,
      colback=ncmetabg,
      colframe=ncmetaedge,
      boxrule=0.35pt,
      arc=12pt,
      left=0.55cm, right=0.55cm, top=0.45cm, bottom=0.4cm,
      interior style={shade, shading angle=315,
        left color=white!96!ncmetabg,
        right color=ncmetablue!4!ncapricot!8!ncmetabg},
      before skip=0pt, after skip=0.4em,
      grow to left by=1.5pt, grow to right by=1.5pt,
    ]
      \centering
      {\LARGE\bf \@title\par}%
      \def\And{\end{tabular}\hfil\linebreak[0]\hfil\begin{tabular}[t]{c}\bf\rule{\z@}{24\p@}\ignorespaces}%
      \def\AND{\end{tabular}\hfil\linebreak[4]\hfil\begin{tabular}[t]{c}\bf\rule{\z@}{24\p@}\ignorespaces}%
      \begin{tabular}[t]{c}\bf\rule{\z@}{24\p@}\@author\end{tabular}\par
      \vskip 0.12in
      {\small
        \textcolor{black}{\faGithub}\enspace\textbf{Code:}\enspace
        \href{https://github.com/mbzuai-oryx/CVPD}{\textcolor{ncmetablue}{\texttt{https://github.com/mbzuai-oryx/CVPD}}}\\[3pt]
        \textcolor{cyan!60!blue}{\faGlobeAmericas}\enspace\textbf{Project Page:}\enspace
        \href{https://mbzuai-oryx.github.io/CVPD/}{\textcolor{ncmetablue}{\texttt{https://mbzuai-oryx.github.io/CVPD/}}}\\[3pt]
        \raisebox{-0.15em}{\includegraphics[height=1em]{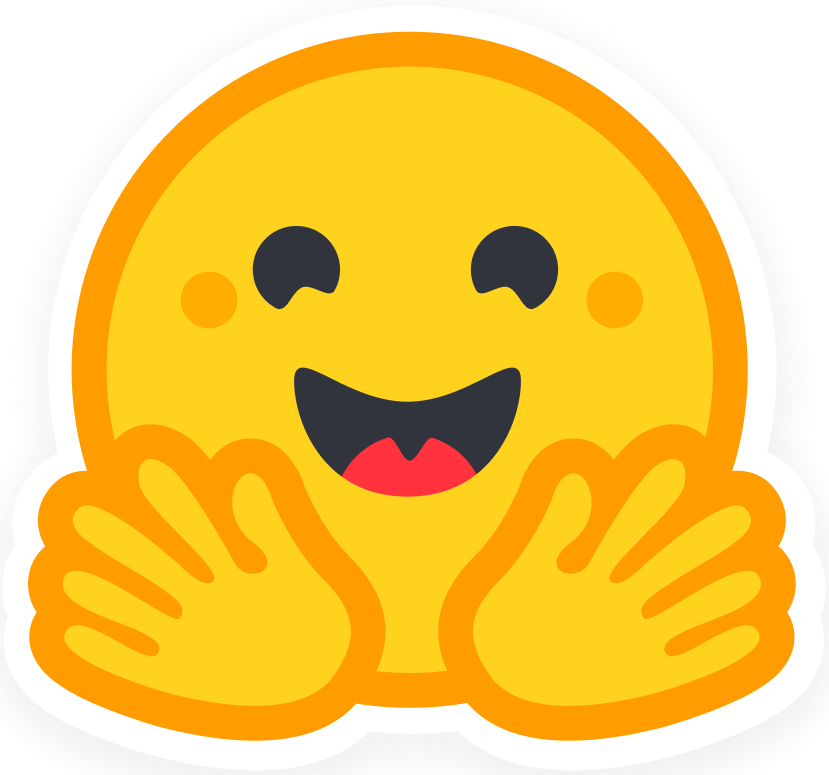}}\enspace\textbf{Model:}\enspace
        \href{https://huggingface.co/shravvvv/CVPD}{\textcolor{ncmetablue}{\texttt{https://huggingface.co/shravvvv/CVPD}}}
      \par}
      \vskip 0.16in
      \begingroup
        \leftskip=1.5em \rightskip=1.5em
        \centerline{\large\bf Abstract}\vspace{0.8ex}
        \small #1\par
      \endgroup
    \end{tcolorbox}%
    \@thanks
    \@notice
  \endgroup
  \let\maketitle\relax
  \let\thanks\relax
}
\makeatother

\begin{document}

\maketitleboxed{Self-improvement for multimodal large language models (MLLMs) is typically driven by reward-based methods that provide only coarse scalar feedback. Distillation offers a richer alternative through dense token-level supervision, but in the visual domain it usually depends on privileged context constructed using external annotations and tools, or stronger models. We introduce \textbf{CVPD} (Contrastive Counterfactual Visual Process Distillation), which, to the best of our knowledge, is the first fully self-contained framework for dense, on-policy, token-level visual self-distillation for MLLMs. CVPD identifies visual blind spots where zooming into a region changes and sharpens the model's answer distribution, while removing the same region leaves the full-image behavior largely unchanged. Such regions reveal perceptual information that the model can encode but fails to consistently utilize under full-image conditioning. We propose a three-gate Counterfactual Criterion that identifies these regions directly from the model's own responses and converts them into dense contrastive supervision for self-distillation. On Qwen3-VL-8B-Instruct, CVPD outperforms six self-evolving baselines across twelve benchmarks, including methods that rely on external GPT-4o supervision, without a single regression. It achieves gains of $+3.60$ on OCRBench, $+3.38$ on MMStar Fine-Grained Perception, and $+3.08$ on MMStar Logical Reasoning, while maintaining or improving performance on broader multimodal benchmarks.
}

\section{Introduction}
\label{sec:intro}

Self-improvement has become a central post-training objective for multimodal large language models (MLLMs), and two broad approaches have emerged. Reward- and correctness-based methods train models on their own generated responses, selecting examples using verifiable outcomes or preference signals~\citep{zelikman2022star,shao2024deepseekmath,thawakar2026evolmm,vise,asg}. These methods avoid human annotation, but the training signal remains coarse. An entire rollout or trajectory receives a scalar reward or advantage rather than dense token-level feedback. Distillation-based methods provide a richer alternative by matching a teacher's next-token distributions on trajectories generated by the student~\citep{agarwal2024gkd,zhao2026opsd,hubotter2026sdpo}. This supplies token-level corrective supervision that scalar rewards cannot provide.

The challenge is that distillation requires privileged context, namely a version of the input that induces a meaningful behavior gap between teacher and student. Historically, constructing this context has relied on information external to the model. In text domains, the problem is relatively straightforward. Appending a reference solution, compiler error, or feedback trace to the prompt is often sufficient to create the desired gap. In vision, however, providing the model with a better perceptual view of the same image is considerably harder. Doing so typically requires external recognition systems, segmentation models, or stronger annotators. This dependence on external tools has remained a major obstacle to fully self-contained dense self-distillation for visual MLLMs.

\begin{figure}[t]
  \centering
  \includegraphics[width=\linewidth]{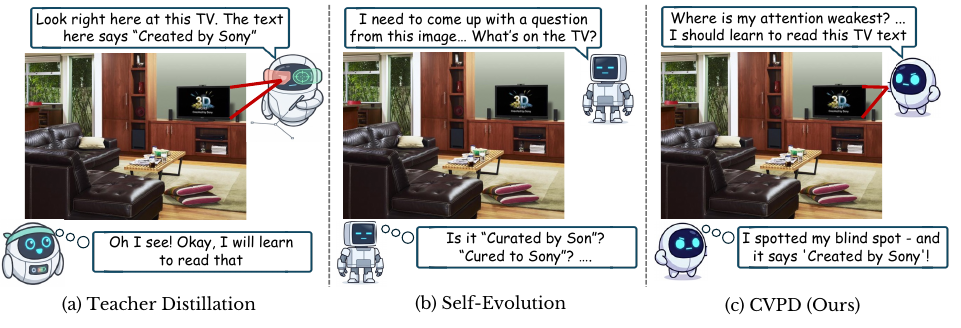}
  \vspace{-7mm}
\caption{\textbf{CVPD turns visual blind spots into dense self-supervision.}
Reward-based self-evolution can identify better or worse responses, but it does
not provide token-level corrective distributions. Standard visual distillation
can provide dense supervision, but typically relies on externally constructed
privileged views. CVPD instead uses the model's own responses to identify
regions where a crop sharpens the answer distribution while a ghosted image
preserves the model's default behavior. The crop view serves as a positive
teacher and the ghost view serves as a counterfactual negative teacher, enabling
dense visual self-distillation without external region proposals, labels, or
stronger annotators.}
  \label{fig:teaser}
  \vspace{-12pt}
\end{figure}

Prior work has approached this challenge from both directions without fully
resolving it. On the distillation side, Vision-OPD~\citep{yuan2026visionopd}
shows that the same MLLM can often answer a fine-grained question correctly
when given an evidence-centered crop while failing on the full image. It
leverages this regional-to-global gap for on-policy visual distillation and
achieves strong results. However, the approach depends on external annotation
and segmentation models to identify evidence regions, generate crop-answerable
questions, and provide consensus answer labels. As a result, the supervision is
dense and on-policy, but the privileged context remains externally constructed.
On the self-improvement side, methods that remove external annotation by
training on reward-filtered or preference-ranked self-generated
data~\citep{zelikman2022star,yuan2024selfrewarding,shao2024deepseekmath}
eliminate this dependency but revert to a much coarser learning signal.
Responses or trajectories receive scalar scores rather than token-level
corrective supervision. Consequently, dense, on-policy, token-level
self-distillation for visual perception without externally constructed
privileged context remains an open problem.

To address this gap, we introduce \textbf{CVPD} (Contrastive Counterfactual
Visual Process Distillation), a framework that recovers privileged visual
context directly from the model's own counterfactual behavior. Our key insight
is that a region provides useful privileged context when zooming into it shifts
and sharpens the model's answer distribution, while removing the same region
does not. This indicates that the model already possesses the relevant
perceptual capability but fails to fully utilize it under full-image
conditioning. CVPD identifies such regions entirely from the model's own
responses and uses them for contrastive training. The crop-conditioned view
acts as a positive teacher, guiding the student toward its latent perceptual
capacity, while the ghost-conditioned view acts as a negative teacher that
captures the model's inattentive default behavior. Together, these signals
provide dense supervision at every token position of the student's own rollout
(Figure~\ref{fig:teaser}). By deriving privileged context entirely from the
model's own behavior, CVPD enables self-contained dense visual
self-distillation without external annotations, tools, or stronger models, as shown in Figure~\ref{fig:teaser}.

Our contributions are summarized as follows:

\begin{itemize}
  \item
    We introduce a three-gate Counterfactual Blind-Spot Criterion that
    identifies image regions containing perceptual information the model can
    use but fails to exploit under full-image conditioning. The criterion
    relies solely on the model's own counterfactual responses, enabling
    privileged context construction without external tools, annotations, or
    stronger models.

  \item
    We develop a Contrastive Self-Distillation objective that converts each
    discovered region into paired positive and negative teacher signals derived
    from the same backbone. This provides dense, per-token, on-policy
    supervision without requiring external rewards, verifiers, or annotators.

  \item
    We show that the proposed criterion reliably identifies regions where crop
    and ghost signals are maximally opposed, and that CVPD delivers consistent
    improvements across twelve visual perception and reasoning benchmarks.
    The largest gains occur on tasks that require precise localized attention,
    while general multimodal capabilities are maintained or improved.
\end{itemize}

\section{Related Work}
\label{sec:related_work}

\noindent\textbf{Self-Improvement for Multimodal LLMs.}
Self-improvement for large language models (LLMs) and multimodal LLMs (MLLMs) is commonly framed as post-training from model-generated data, verifiable outcomes, or intrinsic consistency. Reinforcement learning with verifiable rewards is a widely adopted route. GRPO~\cite{shao2024deepseekmath} removes PPO's learned critic by estimating advantages from groups of sampled responses, and DAPO~\cite{yu2025dapo} refines this objective for large-scale reasoning RL\@. In the multimodal setting, R1-style methods extend this paradigm with rule-based, step-wise, or shared rewards for visual reasoning, as in VLM-R1~\cite{shen2025vlmr1}, R1-VL~\cite{zhang2025r1vl}, and R1-ShareVL~\cite{yao2025r1sharevl}. These methods make reward-based post-training practical, but their feedback is still mediated through rewards and relative advantages rather than dense next-token distributions. Even when rewards are reliable, they indicate which rollout or reasoning step is preferable, not how the model's next-token distribution should change at each visited state.

Self-evolving systems reduce annotation dependence by closing the loop between task generation, solving, and evaluation. In text-only settings, bootstrapping methods such as STaR~\cite{zelikman2022star}, ReST~\cite{gulcehre2023reinforced}, and Self-Rewarding Language Models~\cite{yuan2024selfrewarding} train on model-generated rationales, filtered samples, or self-judged preferences. Recent MLLM variants instantiate this loop through proposer--solver or questioner--solver agents with continuous or intrinsic rewards~\cite{thawakar2026evolmm,sunil2026ireasoner,xu2026rise}, self-consistency and judge-based trajectory reweighting~\cite{wu2026selfjudge}, strategic visual self-play~\cite{wang2025visionzero}, and synthetic or actively retrieved visual environments~\cite{li2026mmzero,he2026activezero}. These works demonstrate the feasibility of improving MLLMs under limited human supervision. However, the training signal remains aggregate. Responses, trajectories, questions, or agent roles receive rewards, scores, or advantages rather than token-level supervision. This motivates the complementary problem of obtaining token-level corrective feedback for visual perception without relying on an external teacher or verifier.

\noindent\textbf{On-Policy Self-Distillation.}
Knowledge distillation~\cite{hinton2015distilling} offers a different supervision regime in which the student matches a teacher's next-token distribution rather than optimizing only terminal or trajectory-level rewards. For autoregressive models, ordinary offline distillation suffers from exposure bias because the student is trained on teacher- or dataset-induced prefixes. Generalized Knowledge Distillation (GKD)~\cite{agarwal2024gkd} addresses this mismatch by querying the teacher on student-generated trajectories, turning distillation into an on-policy objective. More recent on-policy self-distillation methods show that the teacher need not be a separate external model. OPSD~\cite{zhao2026opsd} conditions the same LLM on privileged reasoning traces or reference solutions, while SDPO~\cite{hubotter2026sdpo} conditions the model on rich textual feedback and distills the resulting feedback-aware predictions back into the unconditioned policy. Further variants calibrate or localize this dense signal, for example through outcome-guided logit steering~\cite{yang2026ogls}, reflection-guided error-localized distillation~\cite{zhao2026rosd}, and contrastive evidence policy optimization~\cite{heakl2026cepo}. The common principle is that a context-induced behavior gap within the same model can be converted into dense token-level supervision on the student's own sampled prefixes.

The vision setting lacks an equally direct source of privileged context. A reference solution, compiler error, or feedback message can be appended to an LLM prompt, but giving an MLLM a better perceptual view of the same image requires constructing that view. Vision-OPD~\cite{yuan2026visionopd} identifies the regional-to-global gap: the same MLLM may answer a fine-grained question correctly from an evidence-centered crop while failing on the full image. It uses this gap to distill a crop-conditioned teacher into a full-image student with token-level on-policy divergence, and closely related concurrent work studies visual-advantage weighting for VLM distillation~\cite{liu2026vaopd}. Vision-OPD is the closest prior work to ours. However, its regional data synthesis is not derived solely from the target model's own behavior. Object-recognition and segmentation systems propose evidence regions, an MLLM generates crop-answerable questions, and Qwen3.5-397B provides consensus answer labels used for filtering and alternative training baselines. Thus, although the teacher distribution used during distillation comes from the same backbone, the crop-question supervision is externally constructed rather than self-discovered from the target model's counterfactual behavior.

This leaves open whether dense visual self-distillation can be achieved without externally constructed privileged context. Existing methods either rely on reward-based supervision, which provides only coarse training signals, or on dense distillation objectives whose privileged context is produced through external data synthesis, annotation pipelines, or stronger models. As a result, dense token-level supervision and fully self-contained self-improvement have remained largely separate directions in visual MLLM training. We address this gap by deriving privileged visual context directly from the model's own behavior rather than from externally generated regions, questions, labels, or verifiers. In doing so, our work combines the dense token-level supervision of on-policy self-distillation with a fully self-contained training process, enabling visual self-improvement without external annotations, reward signals, segmentation systems, or stronger annotators.

\section{Preliminaries}
\label{sec:prelim}

On-policy distillation trains a student $p_S$ to match a privileged teacher
$p_T$ along the student's own generated trajectories, avoiding the
prefix-distribution mismatch of offline distillation~\citep{agarwal2024gkd}.
At inference, the student conditions on its own prefixes; training on
teacher-induced prefixes instead introduces a state-distribution shift that
can compound errors over long generation horizons~\citep{ross2011noregret}.
For a multimodal input $(I, q)$ and a rollout $y \sim p_S(\cdot \mid I, q)$,
the objective is
\begin{equation}
  \mathcal{L}_{\mathrm{OPD}}(\theta)
  = \mathbb{E}_{(I,q)\sim\mathcal{D},\;y\sim p_S(\cdot\mid I,q)}
    \!\left[
      \frac{1}{|y|}\sum_{t=1}^{|y|}
      D\!\left(
        p_T\!\left(\cdot\mid\tilde{I},q,y_{<t}\right)
        \;\Big\Vert\;
        p_S\!\left(\cdot\mid I,q,y_{<t}\right)
      \right)
    \right],
  \label{eq:opd}
\end{equation}
where $\tilde{I}$ is a privileged visual context available to the teacher but
withheld from the student at inference, and $D$ is a per-token divergence.
Unlike reinforcement-learning approaches that optimize scalar rewards,
on-policy distillation provides dense token-level supervision along the
student's own trajectory.

The effectiveness of Eq.~\eqref{eq:opd} depends on the existence of a
meaningful asymmetry between teacher and student. In self-distillation
settings, where both share the same backbone, this asymmetry must arise from
the conditioning context rather than model capacity. For visual tasks, prior
work has shown that the same MLLM can answer a question correctly from an
evidence-centered crop while failing on the full image~\citep{yuan2026visionopd},
revealing a regional-to-global perception gap that can serve as privileged
context. However, existing approaches construct such context using external
segmentation pipelines, region proposals, or stronger models. Constructing a
privileged visual context directly from the model's own behavior, without any
external supervision, is the central problem addressed by CVPD.

Throughout the paper, we use $D = D_{\mathrm{JS}}$, approximated over the
union of the top-$K$ logit indices from teacher and student at each token
position~\citep{yuan2026visionopd}, with $K{=}100$.

\section{Method: CVPD}
\label{sec:cvpd}

\begin{figure}[t]
  \centering
  \includegraphics[width=\linewidth]{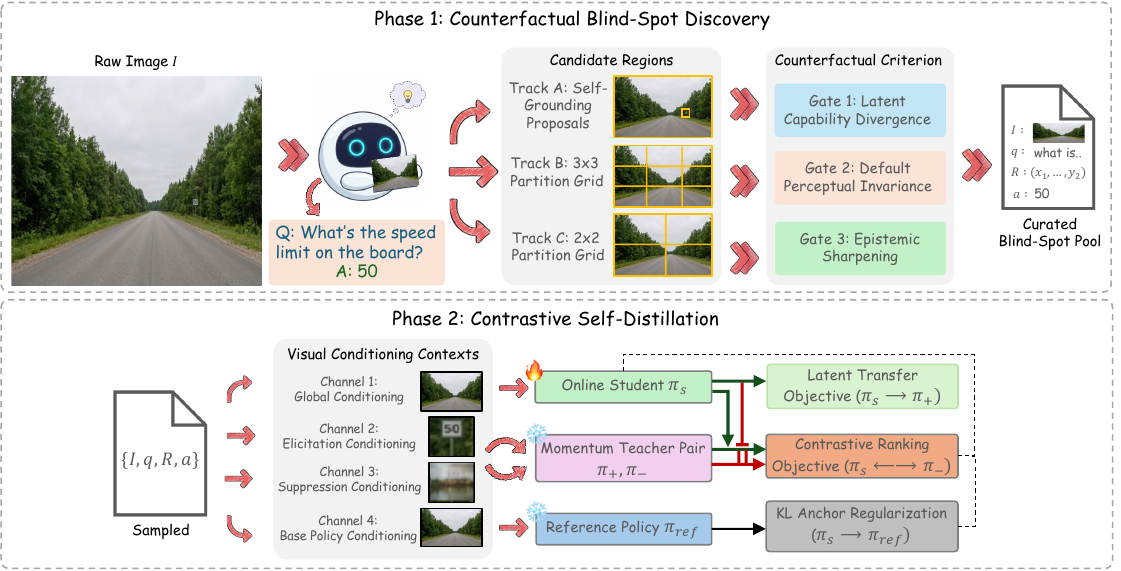}
\caption{\textbf{Overview of CVPD.} CVPD derives all supervision from the model's own responses over unlabeled images, without external labels, annotation tools, segmentation models, or stronger annotators. \textbf{Phase~1 (Counterfactual Blind-Spot Discovery):} The model generates a fine-grained question--answer pair $(q,a)$ for each image and proposes candidate regions through self-grounding (Track~A), a $3{\times}3$ grid (Track~B), and a $2{\times}2$ grid (Track~C). The three-gate Counterfactual Criterion (Eqs.~\eqref{eq:g1}--\eqref{eq:g3}) selects regions where a crop changes and sharpens the answer distribution while a ghosted image preserves the original behavior, producing curated tuples $(I,q,R,a)$. \textbf{Phase~2 (Contrastive Self-Distillation):} Each tuple instantiates four policies from the same backbone under different visual views. The online student $\pi_s$ learns from a crop-conditioned teacher $\pi_+$, a ghost-conditioned teacher $\pi_-$, and a frozen reference policy $\pi_{\mathrm{ref}}$. Their latent-transfer, contrastive-ranking, and KL-anchor objectives together form $\mathcal{L}_{\mathrm{CVPD}}$ (Eq.~\eqref{eq:cvpd_obj}).}
  \label{fig:pipeline}
\end{figure}

Fully unsupervised on-policy self-distillation requires a region where the
model underuses its own perceptual capacity, together with a way to locate and
verify that gap using only the model itself. CVPD identifies such regions
through counterfactual probing. A region is treated as a blind spot when
zooming into it changes and sharpens the model's answer distribution, while
erasing it leaves the full-image behavior largely unchanged. This three-gate
criterion (Section~\ref{sec:cvpd:blindspot}) verifies that a meaningful
teacher--student gap exists and makes the ghost-conditioned view a principled
proxy for the model's inattentive default. CVPD then trains with two teachers:
a crop-conditioned positive teacher that pulls the student toward sharper
regional perception, and a ghost-conditioned negative teacher that pushes it
away from its measured default behavior at every token position. By deriving
both signals from the model's own counterfactual behavior, CVPD enables
self-contained, contrastive, on-policy self-distillation for fine-grained
visual perception in MLLMs.

\subsection{Self-discovered visual blind spots}
\label{sec:cvpd:blindspot}

Let $I$ be an image, $q$ a question generated by the model itself, and $R$ a
candidate region specified by an axis-aligned bounding box. We define three
views using the same model $p_\theta$. The \emph{full view}
$p_{\mathrm{full}} = p_\theta(\cdot \mid I, q)$ conditions on the unmodified
image. The \emph{crop view}
$p_{\mathrm{crop}} = p_\theta(\cdot \mid \mathrm{crop}(I,R),\,q)$ conditions
on $R$, extracted with a 20\% margin and upscaled to the original resolution.
The \emph{ghost view}
$p_{\mathrm{ghost}} = p_\theta(\cdot \mid \mathrm{ghost}(I,R),\,q)$ conditions
on the full image after replacing pixels inside $R$ with their Gaussian blur.
Each divergence is computed token by token over a short probe answer $a$, using
the top-$K$ union approximation of $D_{\mathrm{JS}}$ with $K{=}100$.

We call $R$ a \textbf{visual blind spot} for $(I, q)$ when it satisfies three
counterfactual conditions:
\begin{align}
  \text{(G1)}\;\text{Latent Capability Divergence:}  \quad &
    D_{\mathrm{JS}}(p_{\mathrm{crop}}  \Vert p_{\mathrm{full}}) \geq \tau_{\mathrm{crop}},
  \label{eq:g1}\\[2pt]
  \text{(G2)}\;\text{Default Perceptual Invariance:} \quad &
    D_{\mathrm{JS}}(p_{\mathrm{ghost}} \Vert p_{\mathrm{full}}) \leq \tau_{\mathrm{ghost}},
  \label{eq:g2}\\[2pt]
  \text{(G3)}\;\text{Epistemic Sharpening:}          \quad &
    H[p_{\mathrm{crop}}(y_0)] < H[p_{\mathrm{full}}(y_0)].
  \label{eq:g3}
\end{align}
Gate~(G1) requires the crop view to move the answer distribution away from the
full-image baseline, ensuring that the crop-conditioned teacher carries
information the student does not already express. Gate~(G2) requires that
suppressing the region does not substantially alter the full-image
distribution. This confirms that the model is not using $R$ under full-image
conditioning and makes $p_{\mathrm{ghost}}$ a reliable proxy for its
inattentive default behavior. Gate~(G3) requires the crop to sharpen the model's
prediction rather than merely perturb it, filtering out regions that destroy
useful context or increase uncertainty. Together, the gates select regions
where the crop and ghost views provide opposing signals: the crop captures what
the model can perceive when it attends to the right region, while the ghost
captures what it defaults to when it does not.

When multiple regions pass all three gates, we rank them by
\begin{equation}
  \mathrm{score}(R)
  = D_{\mathrm{JS}}(p_{\mathrm{crop}} \Vert p_{\mathrm{full}})
  - D_{\mathrm{JS}}(p_{\mathrm{ghost}} \Vert p_{\mathrm{full}})
  + \bigl(H[p_{\mathrm{full}}(y_0)] - H[p_{\mathrm{crop}}(y_0)]\bigr),
  \label{eq:score}
\end{equation}
which rewards large crop-side divergence, low ghost-side perturbation, and a
large entropy reduction under the crop. We retain the highest-scoring region
per image.

\begin{figure}[t]
  \centering
  \includegraphics[width=\linewidth]{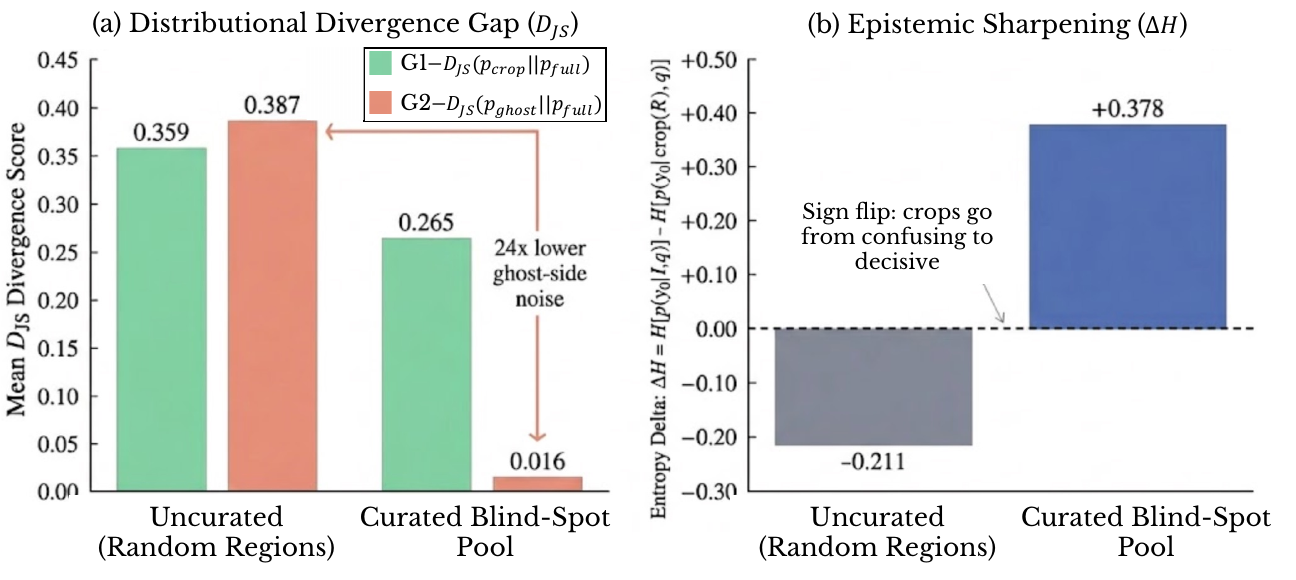}
  \vspace{-7mm}
\caption{\textbf{Distributional signature of counterfactual blind-spot curation.}
We sample $6{,}791$ candidate regions from $500$ raw
images. \textbf{(a)} Uncurated regions produce similar crop/full and ghost/full
divergences ($0.359$ vs. $0.387$), indicating that both interventions perturb
the model in comparable ways. Regions passing the Counterfactual Criterion
retain strong crop-side divergence ($0.265$) while reducing ghost-side
divergence to $0.016$, a $24{\times}$ suppression of irrelevant perturbation.
\textbf{(b)} Curation also reverses the entropy change from $-0.211$ to
$+0.378$, showing that selected crops sharpen the model's answer distribution
rather than merely changing it. Together, these results show that the
Counterfactual Criterion isolates regions where crop and ghost views provide
strongly separated and informative supervision signals.
}
  \label{fig:gap_quality}
\end{figure}

To isolate the effect of the Counterfactual Criterion before full training,
Figure~\ref{fig:gap_quality} analyzes $6{,}791$ candidate regions sampled from
$500$ raw images. Without curation, arbitrary regions produce high JSDs under
both crop and ghost interventions ($0.359$ and $0.387$) and a negative entropy
delta ($\Delta H = -0.211$). In other words, they perturb the model but do not
sharpen its prediction. Regions passing Eqs.~\eqref{eq:g1}--\eqref{eq:g3}
retain strong crop-side sensitivity ($0.265$), suppress ghost-side noise by
$24{\times}$ to $0.016$, and reverse the entropy delta to $+0.378$. The
crop-versus-ghost gap rises from $-0.028$ to $+0.249$ after curation,
supporting the view that the criterion isolates regions where the two teachers provide
genuine and opposing supervision.

\subsection{Contrastive visual process distillation}
\label{sec:cvpd:method}

CVPD uses the Curated Blind-Spot Pool in two phases, illustrated in
Figure~\ref{fig:pipeline} and summarized in Algorithm~\ref{alg:cvpd}. Phase~1
discovers blind-spot tuples from raw images using the counterfactual criterion,
and Phase~2 trains the online student with paired crop and ghost teachers at
each discovered region.

\begin{algorithm}[t]
\caption{CVPD: Contrastive Counterfactual Visual Process Distillation}
\label{alg:cvpd}
\begin{algorithmic}[1]
\Require Unlabeled image pool $\mathcal{I}$; model $p_\theta$; thresholds
         $\tau_{\mathrm{crop}},\;\tau_{\mathrm{ghost}}$
\Statex \textbf{// Phase 1: Counterfactual Blind-Spot Discovery (offline)}
\State $\mathcal{D}_{\mathrm{CVPD}} \gets \emptyset$
\ForAll{$I \in \mathcal{I}$}
  \State $q \gets p_\theta(I)$;\quad $a \gets p_\theta(I,\,q)$
         \Comment{fine-grained question; deterministic probe answer}
  \State $\mathcal{R} \gets$ Track~A proposals on $(I,q)$
         $\,\cup\,$ Track~B $3{\times}3$ grid
         $\,\cup\,$ Track~C $2{\times}2$ grid,\;
         filtered to area ${\in}\,[1\%,\,50\%]$
  \State $\mathcal{R}^\star \gets
         \{R \in \mathcal{R} :
         \text{Eqs.~\eqref{eq:g1}--\eqref{eq:g3} hold}\}$
         \Comment{Counterfactual Criterion}
  \If{$\mathcal{R}^\star \neq \emptyset$}
    \State $\mathcal{D}_{\mathrm{CVPD}} \gets \mathcal{D}_{\mathrm{CVPD}}
           \cup \bigl\{(I,\;q,\;
           \arg\!\max_{R \in \mathcal{R}^\star}\mathrm{score}(R),\;a)\bigr\}$
           \Comment{Eq.~\eqref{eq:score}}
  \EndIf
\EndFor
\Statex \textbf{// Phase 2: Contrastive Self-Distillation (online)}
\State $\bar\theta \gets \theta$;\quad $\beta_0 \gets 10^{-3}$
\For{step $= 1,2,\dotsc$}
  \State $(I,q,R,a) \sim \mathcal{D}_{\mathrm{CVPD}}$
  \State Compute $\pi_s$ \textbf{(grad)},\;
         $\pi_+,\pi_-$ \textbf{(Momentum Teacher Pair, EMA $\bar\theta$, no grad)},\;
         $\pi_{\mathrm{ref}}$ \textbf{(Reference Policy, no grad)}
  \State $\theta \gets \theta - \eta\,\nabla_\theta\,
         \mathcal{L}_{\mathrm{CVPD}}(\theta)$
         \Comment{Eq.~\eqref{eq:cvpd_obj}}
  \State $\bar\theta \gets (1-\alpha)\,\bar\theta + \alpha\,\theta$;\quad
         adapt $\beta_t$ toward $\kappa{=}0.03$
\EndFor
\end{algorithmic}
\end{algorithm}

\textbf{Counterfactual blind-spot sourcing.}
For each raw image $I$, the model generates a fine-grained question $q$
targeting a small attribute, a count, or a textual element, and then samples a
short deterministic probe answer $a$. The pair $(q, a)$ fixes the
teacher-forcing target for all subsequent forward passes on that image.
Candidate regions are collected from three tracks. Track~A obtains
self-grounding proposals by querying the same model for a bounding box on
$(I, q)$. Track~B partitions $I$ into a $3{\times}3$ uniform grid, and
Track~C uses a coarser $2{\times}2$ grid. Every component of the pipeline,
including question generation, region proposal, answer generation, and region
scoring, uses only the model being trained. This self-contained design avoids
the supervision bottleneck that would be reintroduced by annotation tools,
segmentation models, or larger query models. Candidates covering less than
$1\%$ or more than $50\%$ of the image area are discarded as degenerate. The
Counterfactual Criterion in Eqs.~\eqref{eq:g1}--\eqref{eq:g3} is then applied
to the remaining candidates, and the top-scoring blind spot is retained for
each image. Images with no passing region are skipped.

Applied to $15{,}000$ unlabeled images, the pipeline yields
$|\mathcal{D}_{\mathrm{CVPD}}| \approx 2{,}590$ curated tuples, corresponding
to a $17.2\%$ yield. Of the retained blind spots, $27\%$ come from Track~A
self-grounding proposals, $53\%$ from Track~B $3{\times}3$ grid regions, and
$30\%$ from Track~C $2{\times}2$ grid regions. Since an image may contribute
candidates from multiple tracks under multi-region retention, these percentages
describe track membership rather than a mutually exclusive partition
(see Section~\ref{sec:expts}).

\textbf{Contrastive distillation objective.}
Given the Curated Blind-Spot Pool
$\mathcal{D}_{\mathrm{CVPD}} = \{(I_i, q_i, R_i, a_i)\}_{i=1}^N$, Phase~2
instantiates four policies from the same backbone $p_\theta$ under different
visual conditions. The \emph{online student}
$\pi_s = p_\theta(\cdot \mid I, q)$ observes the full image with gradients
enabled. The \emph{Momentum Teacher Pair} $\pi_+$ and $\pi_-$ share an
exponential moving average (EMA) of the LoRA parameters with coefficient
$\alpha{=}0.05$~\citep{yuan2026visionopd}, with gradients stopped:
$\pi_+ = p_{\bar\theta}(\cdot \mid \mathrm{crop}(I,R),\,q)$ conditions on
the crop view, and $\pi_- = p_{\bar\theta}(\cdot \mid \mathrm{ghost}(I,R),\,q)$
conditions on the ghost view. The \emph{reference policy}
$\pi_{\mathrm{ref}} = p_{\theta_0}(\cdot \mid I, q)$ uses the same full-image
input with the LoRA adapter disabled.

The per-token loss combines three objectives:
\begin{equation}
  \mathcal{L}_t
  =
  \underbrace{D_{\mathrm{JS}}(\pi_+ \Vert \pi_s)}_{\displaystyle\text{latent transfer}}
  +\;
  \lambda_{\mathrm{rank}}\cdot
  \underbrace{%
    \max\!\bigl(0,\;m + D_{\mathrm{JS}}(\pi_+ \Vert \pi_s)
                     - D_{\mathrm{JS}}(\pi_- \Vert \pi_s)\bigr)
  }_{\displaystyle\text{contrastive ranking}}
  +\;
  \beta_t\cdot
  \underbrace{D_{\mathrm{KL}}(\pi_s \Vert \pi_{\mathrm{ref}})}_{\displaystyle\text{KL anchor}},
  \label{eq:loss_t}
\end{equation}
and the training objective averages over rollout tokens:
\begin{equation}
  \mathcal{L}_{\mathrm{CVPD}}(\theta)
  =
  \mathbb{E}_{(I,q,R,a)\sim\mathcal{D}_{\mathrm{CVPD}}}
  \!\left[\frac{1}{|a|}\sum_{t=1}^{|a|} \mathcal{L}_t\right],
  \label{eq:cvpd_obj}
\end{equation}
with margin $m{=}0.1$ and rank weight $\lambda_{\mathrm{rank}}{=}0.5$.

The \emph{Latent Transfer Objective} draws the online student $\pi_s$ toward
the crop-conditioned teacher $\pi_+$ at each token position, transferring the
model's regional perceptual capacity into its full-image policy. The
\emph{Contrastive Ranking Objective} makes the ghost-conditioned teacher
$\pi_-$ an active part of training, rather than only a filtering condition
during discovery. Without this margin loss, the ghost view would be used in
Phase~1 and discarded afterward. With it, gate~(G2)'s guarantee that $\pi_-$
reflects the model's inattentive default becomes a per-token pressure that
pushes $\pi_s$ away from that default throughout the rollout. The \emph{KL Anchor Regularization} constrains $\pi_s$ to remain close to the reference policy $\pi_{\mathrm{ref}}$ on the same full-image distribution, preventing targeted perception updates from eroding general language and multimodal capabilities. The coefficient $\beta_t$ is adapted online to maintain a target KL divergence of $\kappa{=}0.03$.
Because the teacher-forcing target $a$ is shared across all four conditioning
contexts, all policies are evaluated at identical token positions, which makes
the per-position divergences in Eq.~\eqref{eq:loss_t} well-defined.

\section{Experiments}
\label{sec:expts}

\textbf{Implementation Details.}
We instantiate policy $\pi$ as Qwen3-VL-8B-Instruct~\citep{qwen3vl}, with a
LoRA adapter~\citep{hu2022lora} of rank $32$ and $\alpha{=}64$ applied to the
attention and MLP projections of the language model backbone; the vision
encoder is frozen. For blind-spot discovery, we set
$\tau_{\mathrm{crop}}=\tau_{\mathrm{ghost}}=0.05$. Question generation uses
temperature $0.7$, top-$p$ $0.9$, and a maximum of $64$ tokens, while the
probe answer is decoded at temperature $0$ and capped at $24$ tokens. Training
uses the AdamW optimizer~\cite{adamw} with learning rate $2{\times}10^{-5}$,
weight decay $0.01$, and gradient clipping at $1.0$, for one epoch over the
Curated Blind-Spot Pool. The KL anchor coefficient is initialized to
$\beta_0=10^{-3}$ and adapted online to maintain a target KL divergence of
$\kappa{=}0.03$. All images (full, crop, and ghost) are resized to
$448{\times}448$, keeping the number of vision tokens identical across the four
forward passes. All models are trained on $8\times$ AMD MI250X GPUs using
bfloat16 precision.

\noindent\textbf{Training Data.}
CVPD operates on entirely unlabeled images with no captions, bounding boxes,
semantic labels, ground-truth answers, reward signals, or external models are
used at any stage. We draw from two unlabeled pools processed by the offline
Counterfactual Blind-Spot Discovery pipeline. Stage~1 consists of $10{,}000$
natural-scene images. Stage~2 consists of $5{,}000$ images in a $40{:}60$
mix of reasoning-domain content (charts, scientific diagrams, and structured
document figures) and natural scenes. The discovery pipeline processes all
$15{,}000$ images and yields ${\sim}2{,}590$ curated
$(I, q, R, a)$ tuples (17.2\% pass rate through the Counterfactual Criterion).
Of the retained blind-spot regions, 53\% derive from Track~B ($3{\times}3$
partition), 30\% from Track~C ($2{\times}2$ partition), and 27\% from Track~A
self-grounding proposals.

\noindent\textbf{Baselines and Evaluation.}
We compare CVPD against six self-evolving baselines on two base models,
Qwen3-VL-4B-Instruct and Qwen3-VL-8B-Instruct~\cite{qwen3vl}. The baselines
include VisPlay~\cite{he2025visplay}, EvoLMM~\cite{thawakar2026evolmm}, and
iReasoner~\cite{sunil2026ireasoner}, which are fully self-supervised and do
not rely on external rewards, verifiers, or annotated data, as well as
VisionZero~\cite{wang2025visionzero} in its CLEVR, Chart, and Real-World (RW)
variants. Among these, VisionZero-CLEVR is label-free, whereas the Chart and
Real-World variants use GPT-4o~\cite{openai2024gpt4technicalreport} during
dataset construction. All methods are trained on the same base models and
evaluated under identical conditions using the lmms-eval
framework~\cite{lmmseval}, HuggingFace Transformers v4.38, and bfloat16
precision on AMD MI250X GPUs. For each baseline, we follow the training setup
reported in the original paper: VisPlay (10 epochs), EvoLMM (1 epoch),
iReasoner (1 epoch), and all VisionZero variants (100 epochs).

We evaluate on twelve benchmarks. Five directly target the fine-grained visual
grounding capabilities that CVPD is designed to improve: OCRBench~\cite{ocrbench}
(localized text understanding), InfoVQA~\cite{infovqa} (information extraction
from dense infographics), and three MMStar~\cite{chen2024mmstar} subscales:
Fine-Grained Perception, Instance Reasoning, and Logical Reasoning. These tasks
require identifying, extracting, and reasoning over specific image regions
rather than relying on holistic scene understanding.

The remaining seven benchmarks evaluate broader multimodal capabilities and
serve as a check that CVPD's targeted perception improvements do not come at
the expense of general performance: ScienceQA~\cite{scienceqa}, RealWorldQA~\cite{realworldqa},
CV-Bench~\cite{cvbench}, AI2D~\cite{ai2d}, MMBench-EN~\cite{mmbench},
MME-Perception~\cite{mme}, and SEED-Image~\cite{seedbench}. Together, these
benchmarks cover scientific reasoning, real-world visual understanding,
diagram interpretation, vision-centric reasoning, holistic perception, and
general multimodal competence.

\subsection{Main Results}
\label{sec:results:main}

\begin{table*}[t]
\centering
\caption{\textbf{Comparison of CVPD against six self-evolving baselines across twelve visual perception and reasoning benchmarks at the 4B and 8B scales.} CVPD achieves the best performance on every benchmark at both scales and is the only method that improves over the base model without regression. The largest gains occur on OCRBench, MMStar Fine-Grained Perception, and MMStar Logical Reasoning, which require precise localized visual attention. Consistent improvements on broader benchmarks indicate that these perception gains do not come at the expense of general multimodal capabilities. Higher is better in all columns. Best result within each scale block is shown in \textbf{bold}.}
\vspace{3mm}
\label{tab:main_results}
\small
\setlength{\tabcolsep}{2.5pt}
\renewcommand{\arraystretch}{1.10}
\resizebox{\textwidth}{!}{%
\begin{tabular}{l|ccccccccc|ccc}
\toprule
\multirow{2}{*}{Method} & \multirow{2}{*}{AI2D} & \multirow{2}{*}{InfoVQA} & \multirow{2}{*}{SciQA} & \multirow{2}{*}{OCR-B} & \multirow{2}{*}{CV-B} & \multirow{2}{*}{RWQA} & \multirow{2}{*}{MMB-EN} & \multirow{2}{*}{MME-P} & \multirow{2}{*}{SEED-I} & \multicolumn{3}{c}{MMStar} \\
\cmidrule(lr){11-13}
& & & & & & & & & & FG & Inst & Log \\
\midrule

\multicolumn{13}{l}{\textit{Qwen3-VL-4B-Instruct}} \\
\midrule
Base                                & 80.10 & 77.73 & 87.51 & 81.70 & 85.45 & 71.24 & 83.51 & 1702.9 & 78.05 & 60.86 & 69.83 & 62.96 \\
VisPlay~\cite{he2025visplay}              & 80.34 & 77.96 & 87.30 & 81.85 & 85.55 & 72.09 & 83.56 & 1706.4 & 77.95 & 61.26 & 70.08 & 63.31 \\
VisionZero-CLEVR~\cite{wang2025visionzero}  & 80.04 & 78.28 & 88.46 & 81.55 & 85.60 & 71.85 & 83.92 & 1704.4 & 77.65 & 61.16 & 69.98 & 63.26 \\
VisionZero-Chart~\cite{wang2025visionzero}  & 80.76 & 78.74 & 88.50 & 82.05 & 85.65 & 71.33 & 83.87 & 1707.9 & 78.00 & 61.01 & 69.93 & 63.16 \\
VisionZero-RW~\cite{wang2025visionzero}     & 80.31 & 77.94 & 87.73 & 81.80 & 85.65 & 72.16 & 83.41 & 1707.9 & 77.80 & 61.46 & 70.18 & 63.51 \\
EvoLMM~\cite{thawakar2026evolmm}            & 81.13 & 78.58 & 88.46 & 81.95 & 85.75 & 72.07 & 84.30 & 1709.4 & 78.05 & 61.51 & 70.28 & 63.56 \\
iReasoner~\cite{sunil2026ireasoner}         & 81.22 & 78.76 & 88.62 & 82.15 & 85.90 & 72.34 & 84.43 & 1711.4 & 78.10 & 61.81 & 70.53 & 63.91 \\

\midrule

\arrayrulecolor{gray!55}
\arrayrulecolor{black}
\rowcolor{cvpdrow}
CVPD (Ours) & \textbf{82.35} & \textbf{79.15} & \textbf{89.05} & \textbf{84.35} & \textbf{87.15} & \textbf{73.45} & \textbf{84.60} & \textbf{1715.5} & \textbf{78.20} & \textbf{63.35} & \textbf{71.20} & \textbf{65.25} \\

\midrule

\multicolumn{13}{l}{\textit{Qwen3-VL-8B-Instruct}} \\
\midrule
Base   & 83.31 & 81.23 & 90.88 & 82.80 & 86.13 & 69.28 & 84.71 & 1716.5 & 78.18 & 60.25 & 73.03 & 61.69 \\
VisPlay~\cite{he2025visplay}              & 83.45 & 81.35 & 91.10 & 82.95 & 86.20 & 69.65 & 84.90 & 1720.5 & 78.55 & 60.70 & 73.35 & 62.10 \\
VisionZero-CLEVR~\cite{wang2025visionzero}  & 83.15 & 81.25 & 90.95 & 82.65 & 86.25 & 69.45 & 84.78 & 1718.0 & 78.25 & 60.55 & 73.25 & 62.00 \\
VisionZero-Chart~\cite{wang2025visionzero}  & 83.75 & 82.55 & 91.00 & 83.15 & 86.30 & 69.45 & 84.85 & 1721.0 & 78.60 & 60.40 & 73.20 & 61.90 \\
VisionZero-RW~\cite{wang2025visionzero}     & 83.35 & 81.25 & 90.95 & 82.90 & 86.30 & 69.75 & 84.95 & 1721.5 & 78.45 & 60.85 & 73.45 & 62.30 \\
EvoLMM~\cite{thawakar2026evolmm}            & 83.70 & 81.55 & 91.45 & 83.05 & 86.35 & 69.70 & 85.05 & 1723.0 & 78.65 & 60.95 & 73.55 & 62.35 \\
iReasoner~\cite{sunil2026ireasoner}         & 83.82 & 81.85 & 91.85 & 83.25 & 86.50 & 69.85 & 85.20 & 1725.0 & 78.72 & 61.25 & 73.80 & 62.70 \\

\midrule
\arrayrulecolor{gray!55}
\arrayrulecolor{black}
\rowcolor{cvpdrow}
CVPD (Ours) & \textbf{84.63} & \textbf{82.82} & \textbf{92.81} & \textbf{86.40} & \textbf{88.27} & \textbf{71.07} & \textbf{86.10} & \textbf{1733.1} & \textbf{78.78} & \textbf{63.63} & \textbf{74.87} & \textbf{64.77} \\

\bottomrule
\end{tabular}%
}
\end{table*}

Table~\ref{tab:main_results} compares CVPD against six self-evolving baselines
across twelve benchmarks at both the 4B and 8B scales. CVPD achieves the best
performance on every benchmark at both scales and is the only method that does
not regress on any benchmark relative to the base model. The gains are largest
on tasks that require precise localization and extraction of visual evidence,
while benchmarks emphasizing global scene understanding and general multimodal
reasoning show smaller but consistent improvements. This pattern aligns with the
design of CVPD, which strengthens the model's ability to utilize visual details
that are often overlooked under standard full-image conditioning.

The largest gains on Qwen3-VL-8B-Instruct are observed on OCRBench
($82.80 \to 86.40$, $+3.60$), MMStar Fine-Grained Perception
($60.25 \to 63.63$, $+3.38$), and MMStar Logical Reasoning
($61.69 \to 64.77$, $+3.08$). OCRBench directly evaluates the ability to
identify and read localized text regions, making it particularly sensitive to
improvements in visual attention. Similar gains are observed on MMStar Instance
Reasoning ($+1.84$) and InfoVQA ($+1.59$), both of which require reasoning over
specific visual evidence within the image. ScienceQA also improves
substantially ($+1.93$), as many questions depend on correctly
interpreting labeled components and spatial relationships in scientific
diagrams. The same trend holds at the 4B scale, where the largest gains again
appear on OCRBench ($+2.65$), MMStar FG ($+2.49$), and MMStar Log ($+2.29$),
indicating that the benefits of CVPD are consistent across model sizes.

CVPD also improves performance across broader benchmarks, including AI2D,
MMBench-EN, MME-Perception, SEED-Image, RealWorldQA, and CV-Bench, without any
loss in general capability. At the 8B scale, gains range from $+0.60$
(SEED-Image) to +1.93 on ScienceQA, alongside a $+16.6$ point improvement on
MME-P. These results are consistent with the role of KL Anchor Regularization
(Eq.~\eqref{eq:loss_t}), which constrains the student to remain close to the
reference policy on full-image inputs while allowing targeted perception
improvements to be incorporated. In contrast, several baselines exhibit
regressions on individual benchmarks. For example, VisionZero-CLEVR falls below
the base model on AI2D and OCRBench at both scales, while VisionZero-RW
regresses on multiple general benchmarks at the 4B scale. CVPD avoids such
trade-offs across the entire evaluation suite.

It is worth noting that VisionZero-Chart and VisionZero-RW rely on GPT-4o
during dataset construction, whereas CVPD derives its training signal entirely
from unlabeled images and the model's own counterfactual responses. Despite
using no external annotator, verifier, or reward model, CVPD outperforms both
methods on every benchmark at both scales. This suggests that dense
self-discovered distributional supervision provides a stronger learning signal
than externally constructed reward-based feedback in this setting.

\subsection{Ablation Study}
\label{sec:results:ablation}

\begin{table*}[t]
\centering
\caption{\textbf{Ablation study of CVPD components on Qwen3-VL-8B-Instruct.}
Each row removes or modifies a single component while holding all others
fixed. Replacing curation with random regions (row d) and removing the
contrastive ranking objective (row a) produce the two largest performance
drops, directly validating the Counterfactual Criterion and the
ghost-conditioned negative teacher as the primary contributors to CVPD's
gains. All values are accuracy~(\%) except MME-P (sum score, 0--2000).
Best result per column in \textbf{bold}.}
\vspace{3mm}
\label{tab:ablation}
\small
\setlength{\tabcolsep}{2.5pt}
\renewcommand{\arraystretch}{1.10}
\resizebox{\textwidth}{!}{%
\begin{tabular}{l|ccccccccc|ccc}
\toprule
\multirow{2}{*}{Configuration}
  & \multirow{2}{*}{AI2D}
  & \multirow{2}{*}{InfoVQA}
  & \multirow{2}{*}{SciQA}
  & \multirow{2}{*}{OCR-B}
  & \multirow{2}{*}{CV-B}
  & \multirow{2}{*}{RWQA}
  & \multirow{2}{*}{MMB-EN}
  & \multirow{2}{*}{MME-P}
  & \multirow{2}{*}{SEED-I}
  & \multicolumn{3}{c}{MMStar} \\
\cmidrule(lr){11-13}
 & & & & & & & & & & FG & Inst & Log \\
\midrule
Base (Qwen3-VL-8B-Instruct)
  & 83.31 & 81.23 & 90.88 & 82.80 & 86.13 & 69.28 & 84.71 & 1716.5 & 78.18
  & 60.25 & 73.03 & 61.69 \\
(a) $\lambda_{\text{rank}}{=}0$ (no contrastive)
  & 83.55 & 81.80 & 91.58 & 84.10 & 86.36 & 69.52 & 84.94 & 1722.5 & 78.52
  & 61.47 & 73.69 & 62.80 \\
(b) frozen teacher ($\alpha{=}0$)
  & 83.64 & 82.02 & 91.85 & 84.60 & 86.45 & 69.67 & 85.06 & 1724.8 & 78.61
  & 61.94 & 73.95 & 63.23 \\
(c) high-quality subset only
  & 83.72 & 82.22 & 92.08 & 85.03 & 86.53 & 69.77 & 85.14 & 1726.8 & 78.68
  & 62.35 & 74.17 & 63.60 \\
(d) no curation (random regions)
  & 83.45 & 81.50 & 91.50 & 83.80 & 86.25 & 69.45 & 84.83 & 1721.0 & 78.45
  & 61.10 & 73.50 & 62.60 \\
(e) unfrozen vision encoder
  & 83.76 & 82.31 & 92.19 & 85.25 & 86.57 & 69.82 & 85.18 & 1727.8 & 78.72
  & 62.55 & 74.28 & 63.78 \\
\midrule
Full CVPD (Ours)
& \textbf{84.63} & \textbf{82.82} & \textbf{92.81} & \textbf{86.40} & \textbf{88.27} & \textbf{71.07} & \textbf{86.10} & \textbf{1733.1} & \textbf{78.78} & \textbf{63.63} & \textbf{74.87} & \textbf{64.77} \\
\bottomrule
\end{tabular}%
}
\end{table*}

Table~\ref{tab:ablation} evaluates the contribution of each component of CVPD
on Qwen3-VL-8B-Instruct by removing or modifying one design choice at a time.
We focus on OCRBench, MMStar Fine-Grained Perception, and MMStar Logical
Reasoning, as these benchmarks exhibit the largest variation across
configurations and are most sensitive to improvements in fine-grained visual
perception.

\noindent\textbf{Counterfactual Criterion (d).}
Replacing the three-gate criterion with randomly selected regions results in the
largest performance drop among all ablations. OCRBench decreases by $2.60$
points ($86.40 \to 83.80$), while MMStar FG drops by $2.53$ points
($63.63 \to 61.10$). This identifies the Counterfactual Criterion as the
primary source of CVPD's gains. Figure~\ref{fig:gap_quality} provides further
evidence: random regions produce nearly identical crop- and ghost-side
divergences ($0.359$ and $0.387$) and a negative entropy delta ($-0.211$),
indicating that arbitrary crops introduce noise rather than informative
supervision. In contrast, the proposed criterion suppresses ghost-side
divergence to $0.016$ while preserving crop-side sensitivity ($0.265$),
yielding a positive entropy delta of $+0.378$. These results prove that the
criterion identifies regions that provide strong and complementary positive and
negative training signals.

\noindent\textbf{Contrastive Ranking Objective (a).}
Setting $\lambda_{\mathrm{rank}} = 0$ removes the ranking loss while retaining
the crop-conditioned teacher. Performance drops substantially, with OCRBench
falling to $84.10$ ($-2.30$) and MMStar FG to $61.47$ ($-2.16$), representing
the second-largest degradation in the table. This shows that learning solely
from the crop-conditioned teacher is insufficient. The additional signal from
the ghost-conditioned teacher provides an important contrastive pressure that
helps the student move away from inattentive behaviors rather than merely
imitating improved ones.

\noindent\textbf{Supporting design choices (b, c, e).}
Removing EMA stabilization (row b, $\alpha = 0$) causes a moderate performance
drop, suggesting that the momentum teacher provides more stable distributional
targets as the student evolves during training. Restricting training to only the
highest-scoring curated tuples (row c) performs better than random region
selection but remains below full CVPD, indicating that broad coverage from the
complete curated set contributes additional learning signal beyond a small set
of highly confident examples. Finally, unfreezing the vision encoder (row e)
slightly reduces performance. This suggests that the pre-trained visual
representations are already well aligned with the language model, and that
updating the encoder using the relatively small curated dataset provides little
benefit while introducing mild instability.

\subsection{Sensitivity Analysis}
\label{sec:results:sensitivity}

\begin{table*}[t]
\centering
\caption{\textbf{Hyperparameter sensitivity of CVPD on Qwen3-VL-8B-Instruct.}
We vary one hyperparameter at a time around the default configuration and report
the metrics most sensitive to the contrastive visual distillation objective:
MMStar fine-grained perception, OCRBench, and MMStar logical reasoning. The
default setting, $\lambda_{\mathrm{rank}}{=}0.5$, $m{=}0.10$, and
$\kappa{=}0.03$, gives the best overall trade-off. Weaker contrastive ranking
under-separates the crop and ghost teachers, while overly strong ranking or KL
relaxation slightly degrades the dense perception signal.}
\vspace{3mm}
\label{tab:hyperparam}
\small
\setlength{\tabcolsep}{5pt}
\renewcommand{\arraystretch}{1.10}
\begin{tabular*}{\textwidth}{@{\extracolsep{\fill}}ll|ccc}
\toprule
Hyperparameter & Value & MMStar FG & OCRBench & MMStar Log \\
\midrule
\multirow{4}{*}{Rank weight $\lambda_{\text{rank}}$}
  & 0.0${}$           & 61.47 & 84.10 & 62.80 \\
  & 0.25${}$          & 62.85 & 85.55 & 64.02 \\
  & 0.5 (default)     & \textbf{63.63} & \textbf{86.40} & \textbf{64.77} \\
  & 1.0${}$           & 63.00 & 85.65 & 64.16 \\
\midrule
\multirow{3}{*}{Margin $m$}
  & 0.05${}$          & 63.10 & 86.15 & 64.50 \\
  & 0.10 (default)    & \textbf{63.63} & \textbf{86.40} & \textbf{64.77} \\
  & 0.20${}$          & 62.85 & 86.00 & 64.30 \\
\midrule
\multirow{3}{*}{KL target $\kappa$}
  & 0.01${}$          & 63.12 & 85.80 & 64.30 \\
  & 0.03 (default)    & \textbf{63.63} & \textbf{86.40} & \textbf{64.77} \\
  & 0.10${}$          & 63.40 & 86.20 & 64.55 \\
\bottomrule
\end{tabular*}
\end{table*}

Table~\ref{tab:hyperparam} evaluates the sensitivity of CVPD to its three
training hyperparameters, varying one parameter at a time while keeping all
others fixed. Results are reported on the three benchmarks most affected by the
ablation study.

\noindent\textbf{Rank weight $\lambda_{\mathrm{rank}}$.}
Performance peaks at the default value of $\lambda_{\mathrm{rank}} = 0.5$ and
degrades slightly when moved in either direction. At $0.25$, performance
remains strong (MMStar FG: $62.85$, OCRBench: $85.55$), indicating that CVPD is
effective across a reasonably broad range of values. Increasing the weight to
$1.0$ leads to a small decline ($63.00$, $85.65$), suggesting that
over-emphasizing the contrastive objective can partially interfere with the
underlying distillation signal. Overall, the method is relatively robust within
the $0.25$--$0.50$ range.

\noindent\textbf{Margin $m$ and KL target $\kappa$.}
Both parameters exhibit limited sensitivity across the tested ranges, with
variations below one point on all three benchmarks. The default settings are
consistently at or near the best-performing configuration. These results
indicate that CVPD does not require precise tuning of either parameter and
remains stable across a broad range of values.

\subsection{Blind-Spot Source Analysis}
\label{sec:results:source}

\begin{table*}[t]
\centering
\caption{\textbf{Candidate region source ablation on Qwen3-VL-8B-Instruct.}
Each row trains CVPD using blind spots derived from a single
region-generation track in isolation. All three tracks individually improve
over the base model, with Track~B ($3{\times}3$ grid) as the strongest
standalone source. Combining all three tracks achieves the best results on
all benchmarks, confirming that self-grounding proposals, coarse-grid crops,
and fine-grid partitions surface complementary blind spots at different
spatial granularities. Best result per column in \textbf{bold}.}
\vspace{3mm}
\label{tab:source_ablation}
\small
\setlength{\tabcolsep}{6pt}
\renewcommand{\arraystretch}{1.10}
\resizebox{0.75\textwidth}{!}{%
\begin{tabular}{l|c|ccc}
\toprule
Source & Samples & MMStar FG & OCRBench & MMStar Log \\
\midrule
Grounding only       & $\sim 699$   & 62.50 & 84.90 & 63.30 \\
$2\times 2$ grid only & $\sim 777$   & 62.80 & 85.30 & 63.80 \\
$3\times 3$ grid only & $\sim 1{,}373$ & 63.20 & 85.95 & 64.45 \\
\midrule
All combined (Ours)                 & $\sim 2{,}590$ & 63.63 & 86.40 & 64.77 \\
\bottomrule
\end{tabular}%
}
\end{table*}

Table~\ref{tab:source_ablation} evaluates the contribution of each candidate
region source by training CVPD using blind spots derived from a single track in
isolation.

All three tracks improve upon the base model (MMStar FG:
$60.25 \to 62.50/62.80/63.20$; OCRBench:
$82.80 \to 84.90/85.30/85.95$), indicating that each source is capable of
identifying useful training regions. Among the individual tracks, Track~B
($3{\times}3$ grid) performs best, reaching $63.20$ on MMStar FG and $85.95$
on OCRBench. A likely explanation is that the finer spatial partitioning
produces more candidate regions per image, increasing the chance of capturing
informative localized evidence.

Combining all three tracks yields the strongest overall performance across every
benchmark ($63.63$, $86.40$, $64.77$), consistently outperforming any
individual source, implying that self-grounding proposals, coarse
$2{\times}2$ crops, and fine $3{\times}3$ partitions capture complementary
regions that are not fully covered by any single strategy. Track~A self-grounding proposals account for only $27\%$ of the
retained blind spots, the smallest share among the three sources. This is
consistent with the intuition behind blind spots: regions that the model fails
to attend to are also less likely to be identified through its own grounding
predictions. In contrast, grid-based partitioning systematically explores the
image and is therefore more likely to uncover regions that the model would
otherwise overlook.


\section{Conclusion}
\label{sec:conclusion}

We introduced CVPD, a fully self-contained framework for dense on-policy visual self-distillation in MLLMs. By identifying visual blind spots directly from the model's own counterfactual responses, CVPD recovers dense supervision without relying on external annotations, segmentation tools, reward signals, or stronger models. Across twelve benchmarks and multiple model scales, CVPD consistently improves over the base model and all self-evolving baselines, achieving the largest gains on tasks that require precise localized visual attention while preserving broader multimodal capabilities. These results suggest that a major bottleneck in fine-grained visual perception is not simply model capacity or the availability of labeled data, but the model's ability to reliably utilize perceptual information it already encodes. More broadly, our findings show that dense supervision for visual self-improvement can emerge from the model's own latent perceptual capabilities, without any external source of supervision. This work takes a step toward fully self-contained visual self-distillation and suggests promising future directions for deriving dense supervision directly from a model's own capabilities.

\section{Acknowledgment}

The computations were enabled by resources provided by LUMI hosted by CSC (Finland) and LUMI consortium, and by Berzelius resource provided by the Knut and Alice Wallenberg Foundation at the NSC.

\bibliographystyle{plain}
\bibliography{main}

\clearpage

\end{document}